\documentclass[conference,final,twocolumn]{IEEEtran}
\usepackage{mathrsfs}
\usepackage{graphicx}
\usepackage{float}
\usepackage{booktabs}
\usepackage{hyperref}
\usepackage{cite}
\usepackage{color}
\usepackage{psfrag}
\usepackage{subfigure}
\usepackage{amssymb}
\usepackage{amsthm}
\usepackage{setspace}
\usepackage{epsfig}
\usepackage{pifont}
\usepackage{amsmath}
\usepackage{array}
\usepackage{multicol}
\usepackage{multirow}
\usepackage{pifont}
\usepackage{indentfirst}
\usepackage{amsfonts}
\usepackage{algorithm}
\usepackage{algorithmicx}
\usepackage{algpseudocode}
\usepackage{fancyhdr}
\usepackage{amscd}
\usepackage{bm}
\usepackage{fancyhdr}
\usepackage{enumerate}
\usepackage{color}

\newtheorem{proposition}{Proposition}
\newtheorem{definition}{Definition}

\newcommand{\tabincell}[2]{\begin{tabular}{@{}#1@{}}#2\end{tabular}}

\IEEEoverridecommandlockouts
\begin{document}
	
	\title{Structure of Deep Neural Networks with a Priori Information in Wireless Tasks}
		
\author{
	\IEEEauthorblockN{Jia Guo and Chenyang Yang}
	\IEEEauthorblockA{Beihang University, Beijing, China\\Email: \{guojia, cyyang\}@buaa.edu.cn}
}
	\maketitle
	
\begin{abstract}
Deep neural networks (DNNs) have been employed for designing wireless networks in many aspects, such as transceiver optimization, resource allocation, and information prediction. Existing works either use fully-connected DNN or the DNNs with specific structures that are designed in other domains. In this paper, we show that \emph{a priori} information widely existed in wireless tasks is permutation invariant. For these tasks, we propose a DNN with special structure, where the weight matrices between layers of the DNN only consist of two smaller sub-matrices. By such way of parameter sharing, the number of model parameters reduces, giving rise to low sample and computational complexity for training a DNN. We take predictive resource allocation as an example to show how the designed DNN can be applied for learning the optimal policy with unsupervised learning. Simulations results validate our analysis and show dramatic gain of the proposed structure in terms of reducing training complexity.

\end{abstract}

\section{Introduction}
Deep neural networks (DNNs) have been introduced to design wireless networks recently in various aspects, ranging from signal detection and channel estimation \cite{ye2017power}, multi-cell coordinated beamforming \cite{alkhateeb2018deep}, inter-cell interference management \cite{sun2017learning}, resource allocation \cite{Mark2019Sig,YW19,VTC18GJ,Guo2018Exploiting}, traffic load prediction \cite{wang2017spatiotemporal}, and uplink/downlink channel calibration \cite{huang2019deep}, \emph{etc}.

These research efforts are motivated by the fact that DNNs can learn an input-output relation thanks to the universal approximation theorem \cite{Hornik1989UnivApprox}. For the tasks of transceiver design and resource allocation, the relation is the dependence of a concerned policy (e.g., beamforming vector or power allocation) on the input parameters (e.g., channel gains). For the tasks of information prediction, the relation is the temporal correlation between historical and future samples of a time series (e.g., traffic load at a base station).

In the literature of solving wireless problems with  DNNs, the considered structure of the DNNs is either the fully-connected (FC)-DNN to illustrate how deep learning can be employed for wireless applications \cite{ye2017power,alkhateeb2018deep,huang2019deep,sun2017learning}, or the combination of several variants of existing DNNs adequate for the specific tasks \cite{wang2017spatiotemporal,VTC18GJ,YW19}. Most of these DNNs are trained with labels, disregard the fact that gathering or generating large number of training samples is cost-prohibitive or time-consuming. While the recently proposed frameworks for a large class of wireless problems, learning to optimize, have leveraged the feature of this type of applications such that the labels for training no longer need to generate \cite{Mark2019Sig,YW19,sun2019pimrc}, the sample and computational complexity is still quite high for complex problems. Even though the computational complexity of off-line training is less of a concern in a static environment, wireless networks often operate in highly dynamic environments. Hence the DNNs have to be re-trained whenever the input parameters change significantly, or even need to be trained in an on-line manner. For example, the DNN in \cite{alkhateeb2018deep} for coordinated beamforming needs to be re-trained in each time duration in the scale of minutes. Therefore, training DNNs efficiently is critical for wireless applications.

A promising way of designing efficient DNN is to introduce adequate structure for specific tasks. For example, convolutional neural network (CNN) is efficient for learning images, and recurrent neural network (RNN) is efficient for predicting time-series. While some wireless applications are analogous to image processing hence can use CNN and some wireless applications are concerned with information prediction hence can use RNN, wireless problems themselves also have unique feature. In many wireless tasks, the relation between concerned solutions and relevant  parameters satisfies a property of permutation invariant. For example, if the permutation of multiple users' channel gains changes, the permutation of allocated resources to the users also changes. This is because each user's resource allocation depends on its own channel but not on the permutation of other users' channels \cite{sun2017learning,alkhateeb2018deep,VTC18GJ,Mark2019Sig,YW19}. Such \emph{a priori information} can be exploited to design the structure of DNNs for these wireless tasks.

%

In this paper, we strive to show how to design a DNN by harnessing the \emph{a priori} information of permutation invariant. We show that for the wireless tasks with such property, the DNN with a special structure can represent the relation of concerned, where the model parameters in most hidden nodes are identical. The proposed DNN with the special structure can be incorporated with any design of DNN that accounting for the features of wireless problems.
To illustrate how to use the DNN structure, we consider a problem of finding the optimal predictive resource allocation (PRA) with unsupervised learning. Simulation results show that much fewer samples and much lower computational complexity are required for training the DNN with parameter sharing. The proposed DNN structure can be widely applied for wireless tasks, including but not limited to the tasks in \cite{sun2017learning,wang2017spatiotemporal,alkhateeb2018deep,VTC18GJ,YW19,Mark2019Sig,sun2019pimrc}.


\emph{Notations}: ${\mathbb E}\{\cdot\}$ denotes mathematical expectation, $\|\cdot\|$ denotes two-norm, $\|\cdot\|_1$ denotes the summation of all the absolute-termed elements in a vector or matrix, and $(\cdot)^{\rm H}$ denotes conjugate transpose, ${\bm 1}$ denotes a column vector with all elements being $1$.
\vspace{-1mm}
\section{DNN for Tasks with Permutation Invariant}\label{sec: param share}
\vspace{-1mm}
For many wireless tasks such as resource allocation and transceiver design, the solution ${\bf y}=f({\bf x})$ is a function of known parameters ${\bf x}$, where ${\bf x}=[{\bf x}^1, \cdots, {\bf x}^K]$ and ${\bf y}=[{\bf y}^1, \cdots, {\bf y}^K]$.
\vspace{-1mm}
\begin{definition} \label{def: 1}
For arbitrary permutation to ${\bf x}$, i.e., $\tilde{\bf x}=[{\bf x}^{N_1},\cdots,{\bf x}^{N_k}]$, where $N_1,\cdots,N_K$ is arbitrary permutation of $1,\cdots,K$, if $\tilde{\bf y}=f(\tilde{\bf x})=[{\bf y}^{N_1},\cdots,{\bf y}^{N_K}]$, which is the corresponding permutation to ${\bf y}=f({\bf x})$, then $f({\bf x})$ is permutation invariant to ${\bf x}$.
\end{definition}
\vspace{-2mm}
A widely-existed relationship between the solution ${\bf y}$ and
the parameters ${\bf x}$ in wireless communications is permutation invariant to ${\bf x}$, where the $k$th block in ${\bf y}$ can be expressed as
\begin{equation} \label{eq: perm inva-ex}
\textstyle {\bf y}^k = \zeta(\psi({\bf x}^k), \sum_{n=1,n\neq k}^K \phi({\bf x}^n)), k=1,\cdots,K,
\end{equation}
where $\zeta(\cdot)$, $\psi(\cdot)$ and $\phi(\cdot)$ are arbitrary functions. This is because for any permutation of ${\bf x}$, $\tilde{\bf x}=[{\bf x}^{N_1},\cdots, {\bf x}^{N_K}]$, the $k$th block of $\tilde{\bf y}$ is $\tilde{\bf y}^k = \zeta(\psi({\bf x}^{N_k}),\sum_{n=1,n\neq N_k}^K \phi({\bf x}^n))={\bf y}^{N_k}$. Hence, the solution corresponding to $\tilde{\bf x}$ is $\tilde{\bf y}=[\tilde{\bf y}^1,\cdots,\tilde{\bf y}^K]=[{\bf y}^{N_1},\cdots,{\bf y}^{N_K}]$.

For example, for the task of optimizing transceivers to coordinate interference among $K$ users, ${\bf x}^k$ is the channel vector of the $k$th user and ${\bf y}^k$ is the transceiver for the $k$th user \cite{sun2017learning}. The optimized transceiver can be expressed as ${\bf y}^k=\zeta\Big(\psi({\bf x}^k)/(\sum_{n=1,n\neq k}^K \phi({\bf x}^n))\Big)$ that depends on the signal to interference ratio, where $\psi({\bf x}^k)$ reflects the received signal at the $k$th user, and $\sum_{n=1,n\neq k}^K \phi({\bf x}^n)$ reflects the interference from other users.

In fact, a more general relationship than \eqref{eq: perm inva-ex} can be proved as permutation invariant to ${\bf x}$.

\begin{proposition}\label{pp: 1}
	The sufficient and necessary condition that ${\bf y}=f({\bf x})$ is permutation invariant to ${\bf x}$ is,
	\begin{equation} \label{eq: perm inva}
	\textstyle {\bf y}^k = \zeta(\psi({\bf x}^k), {\cal F}_{n=1,n\neq k}^K \phi({\bf x}^n)), k=1,\cdots,K,
	\end{equation}
	where $\zeta(\cdot), \psi(\cdot)$ and $\phi(\cdot)$ are arbitrary functions, and ${\cal F}$ is arbitrary operation satisfying the commutative law.
	\begin{IEEEproof}
		See Appendix \ref{appendix: A}.	
	\end{IEEEproof}
\end{proposition}

The relationship in \eqref{eq: perm inva} has two properties: (i) the impact of ${\bf x}^k$ and the impact of other blocks ${\bf x}^n,n\neq k$ in ${\bf x}$ on ${\bf y}^k$ are different, and (ii) the impact of every single block ${\bf x}^n,n\neq k$ on ${\bf y}^k$ does not need to differentiate.

\begin{figure}[!htb]
	\centering
	\vspace{-3mm}
	\includegraphics[width=.7\linewidth]{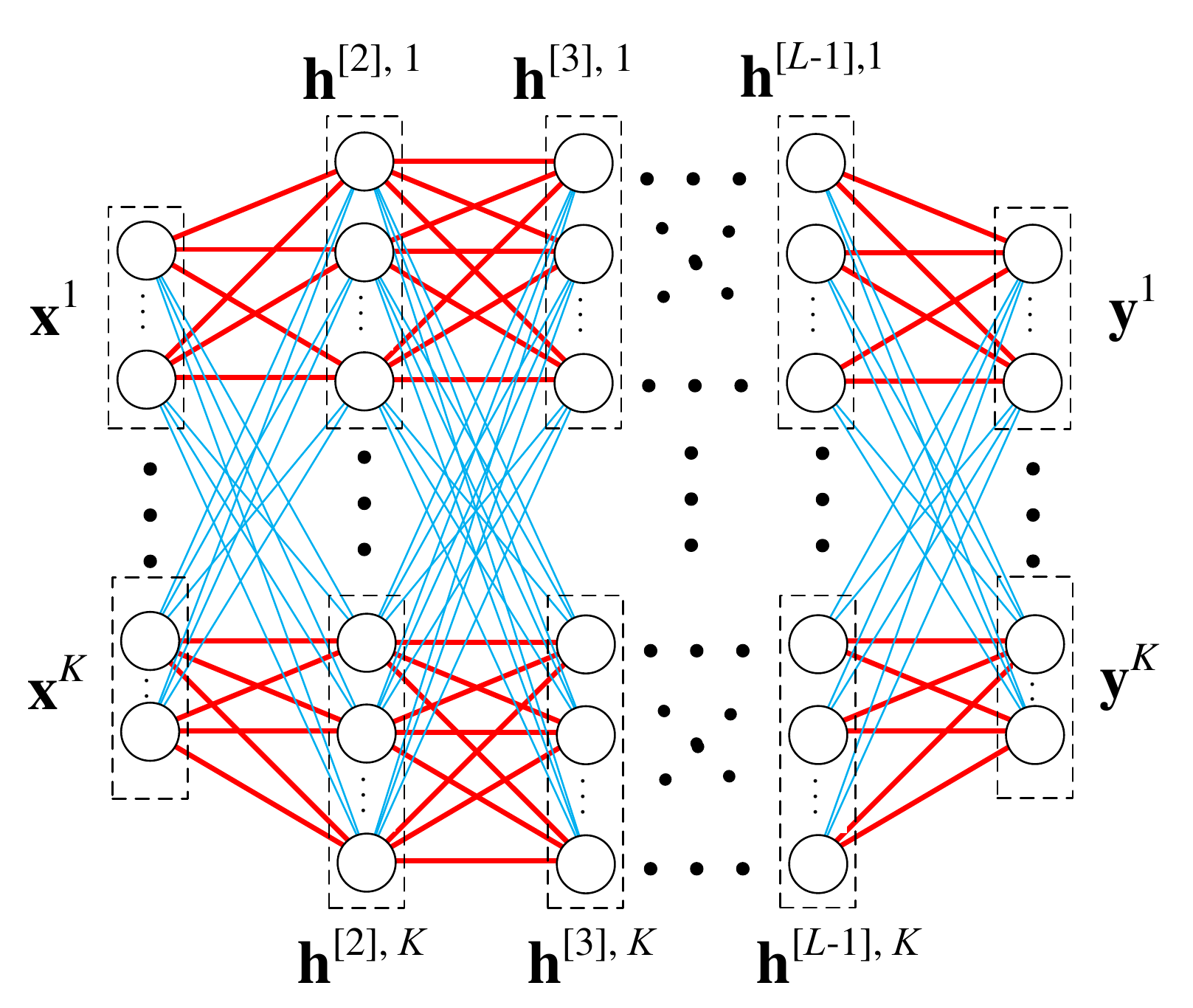}
	\vspace{-2mm}
	\caption{Structure of the DNN with parameter sharing, where in each layer the links with the same color are with the same weights, i.e., ${\bf U}^{[l-1,l]}$ and ${\bf V}^{[l-1,l]}$.}
	\vspace{-2mm}
	\label{fig:dnn}
	\vspace{-1mm}
\end{figure}

When we use DNN for wireless tasks, the essential design goal is to learn a function
${\bf y}=f({\bf x}, {\bf W})$,
where ${\bf x}$ and ${\bf y}$ are respectively the input and output of the DNN, and ${\bf W}$ is the model parameters that need to be trained. When $f({\bf x}, {\bf W})$ is permutation invariant to ${\bf x}$, we can reduce the number of model parameters by introducing parameter sharing into the FC-DNN, as shown in Fig. \ref{fig:dnn}.

The input-output relation of FC-DNN can be expressed as,
\vspace{-3mm}
\begin{equation}\label{eq: FC func}
{\bf y}\!=\!f({\bf x}, {\bf W})\!\triangleq\! g({\bf W}^{[L-1,L]} g(\cdots g({\bf W}^{[1,2]}{\bf x}\!+\!{\bf b}^{[2]})\cdots)\!+\!{\bf b}^{[L]}),
\end{equation}
where $g(\cdot)$ is the activation function of each layer (here we let the activation function in each layer be same for notational simplicity), ${\bf W}^{[l-1,l]}$ is the weight matrix between the $(l-1)$th layer and the $l$th layer, ${\bf b}^{[l]}$ is the bias of the $l$th layer, ${\bf W} = \{\{{\bf W}^{[l-1,l]}\}_{l=2}^L, \{{\bf b}^{[l]}\}_{l=1}^L\}$, and $L$ is the number of hidden layers.

We can construct the weight matrices with special structure to make $f({\bf x}, {\bf W})$ permutation invariant to ${\bf x}$.

\begin{proposition} \label{pp: 2}
The weight matrices with the following structure can make $f({\bf x}, {\bf W})$ in \eqref{eq: FC func} permutation invariant to ${\bf x}$,
\begin{equation}\label{weight mat0}
{\bf W}^{[l-1,l]} = \left[
\begin{tabular}{cccc}
${\bf U}^{[l-1,l]}$ & ${\bf V}^{[l-1,l]}$ & $\cdots$ & ${\bf V}^{[l-1,l]}$ \\
${\bf V}^{[l-1,l]}$ & ${\bf U}^{[l-1,l]}$ & $\cdots$ & ${\bf V}^{[l-1,l]}$ \\
$\vdots$ & $\vdots$ & $\ddots$ & $\vdots$ \\
${\bf V}^{[l-1,l]}$ & ${\bf V}^{[l-1,l]}$ & $\cdots$ & ${\bf U}^{[l-1,l]}$
\end{tabular}
\right],
\end{equation}
where ${\bf U}^{[l-1,l]}$ and ${\bf V}^{[l-1,l]}$ are sub-matrices with the numbers of rows and columns respectively equal to the numbers of elements in ${\bf h}^{[l],k}$ and ${\bf h}^{[l-1],k}$, and ${\bf h}^{[l],k}$ and ${\bf h}^{[l-1],k}$ are respectively the $k$th block in the output of the $l$th and $(l-1)$th hidden layers, $k=1,\cdots,K, l=2,\cdots,L$.

\begin{IEEEproof}
	See Appendix \ref{appendix: B}.
\end{IEEEproof}
\end{proposition}
%

In \eqref{weight mat0}, all the diagonal sub-matrices of ${\bf W}^{[l-1,l]}$ are ${\bf U}^{[l-1,l]}$, which are the model parameters to learn the impact of ${\bf x}^k$ on ${\bf y}^k$. All the other sub-matrices are ${\bf V}^{[l-1,l]}$, which are the model parameters to learn the impact of ${\bf x}^n, n \neq k$ on ${\bf y}^k$. In this way of parameter sharing, the training complexity of the DNN can be reduced since only two sub-matrices need to be trained in each layer.
In what follows, we provide several examples, say information prediction and learning to optimize wireless systems, to explain the role of the two sub-matrices.

\subsubsection{Information Prediction}
We take the traffic load prediction for $K$ base stations (BSs) as an example \cite{wang2017spatiotemporal}. The $k$th block of the input ${\bf x}$ and output ${\bf y}$ of the DNN are respectively ${\bf x}^k$ and ${\bf y}^k$, where ${\bf x}^k$ includes the historical traffic load and the location information of the $k$th BS, ${\bf y}^k$ is the future traffic load of the $k$th BS. ${\bf U}^{[l-1,l]}$ is used to learn the temporal correlation between the past and future traffic load of each BS, and ${\bf V}^{[l-1,l]}$ is used to learn the temporal-spatial correlation between the future traffic load of each BS and the past traffic load of other BSs.

\subsubsection{Interference Coordination}
To learn the optimized transceivers for coordinating inter-cell interference as in \cite{sun2017learning}, ${\bf x}^k$ and ${\bf y}^k$ are the instantaneous channel vector and  transceiver for the $k$th user, respectively. We use ${\bf U}^{[l-1,l]}$ to learn the impact of each user's channel on its own transceiver, and use ${\bf V}^{[l-1,l]}$ to learn the impact of other users' channels on each user's transceiver, i.e., interference.

\subsubsection{Predictive Resource Allocation}
To learn the optimal resource allocation plan for each user as in \cite{VTC18GJ}, ${\bf x}^k$ is the predicted average data rates of the $k$th user in a prediction window with duration $T_f$, and ${\bf y}^k$ is the predicted plan for the user. We use ${\bf U}^{[l-1,l]}$ to learn the impact of each user's past average data rates on its own plan, and use ${\bf V}^{[l-1,l]}$ to learn the impact of other users's average rates on each user's plan, i.e., the total resource occupied by other users.

When $T_f=1$, PRA degenerates to non-PRA. Since the dimension of ${\bf x}^k$ and ${\bf y}^k$ and hence the scale of the DNN increases with $T_f$, in the sequel we use PRA to illustrate how to use parameter sharing to reduce the training complexity.
\section{Case Study: Predictive Resource Allocation}
In this section, we illustrate how the optimal solution of a resource allocation policy can be learned by a DNN with parameter sharing. We consider unsupervised learning, while the basic idea is also applicable to supervised learning.
\subsection{Problem Statement and Formulation}
\subsubsection{System Model}
Consider a wireless network with $N_{\rm b}$ cells, where each BS is equipped with $N_{\rm tx}$ antennas and connected to a central processor (CP). The BSs may serve both real-time traffic (e.g., phone calls) and non-real-time (NRT) traffic (e.g., file downloading). Since real-time service is with higher priority, NRT traffic is served with residual resources of the network after the quality of real-time service is guaranteed.

We design PRA for mobile stations (MSs) requesting NRT service. Suppose that $K \leq K_{\max}$ MSs in the network initiate requests at the beginning of a prediction window, and the $k$th MS (denoted as MS$_k$) requests a file with $B_k$ bits, where $K_{\max}$ is the maximal number of MSs that the network can serve simultaneously.

Time is discretized into frames each with duration $\Delta$, and each frame includes $T_s$ time slots each with duration of unit time. The durations are defined according to the channel variation, i.e., the coherence time of large scale fading (i.e., path-loss and shadowing) and small scale fading due to user mobility.
The prediction window contains $T_f$ frames.

Assume that a MS is only associated to the BS with the highest average channel gain (i.e.,  large scale channel gain) in each frame. To avoid multi-user interference, each BS serves only one MS with all residual bandwidth and transmit power after serving real-time traffic in each time slot, and serves multiple MSs in the same cell in different time slots. Assume that the residual transmit power is proportional to the residual bandwidth \cite{YTCOM2016}, then the achievable rate of  MS$_k$ in the $t$th time slot of the $j$th frame can be expressed as $R_{j,t}^k =  W_{j,t}\log_2\Big(1+\frac{\alpha_j^k \|{\bm \gamma}_{j,t}^k\|^2}{\sigma_0^2} P_{\max}\Big)$, where $W_{j,t}$ and $P_{\max}$ are respectively the residual bandwidth in the $t$th time slot of the $j$th frame and the maximal transmit power, $\sigma_0^2$ is the noise power, ${\bm \gamma}^{k}_{j,t} \in \mathbb{C}^{N_{\rm tx} \times 1}$ is the small scale channel vector with ${\mathbb E}\{{\bm \gamma}^{k}_{j,t}\}=N_{\rm tx}$, $\alpha_j^k$ is the large scale channel gain. When $N_{\rm tx}$ and $T_s$ are large, it is easy to show that the time-average rate in the $j$th frame of MS$_k$ can be accurately approximated as,
\begin{eqnarray}\label{R}	
    R_j^k  \!&\!\!\!\triangleq \!\!&\! \frac{1}{T_s}\sum_{t=1}^{T_s} R_{j,t}^k \!=\! \frac{1}{T_s}\sum_{t=1}^{T_s} W_{j,t}\log_2\Big(1+\frac{\alpha_j^k \|{\bm \gamma}_{j,t}^k\|^2}{\sigma_0^2}P_{\max}\Big)\notag\\
    \! &\!\!\! \approx \!\!\!&\!  W_j\log_2\Big(1+\frac{\alpha_j^k N_{\rm tx}}{\sigma_0^2}P_{\max}\Big),
\end{eqnarray}
where $W_j=\frac{1}{T_s}\sum_{t=1}^{T_s}W_{j,t}$ is the time-average residual bandwidth in the $j$th frame.

The time-average rates of each MS in the frames of the prediction window can either be predicted directly \cite{NB2018} or indirectly by first predicting the trajectory of each MS \cite{LSTMtrjactory17} and the RT traffic load of each BS \cite{wang2017spatiotemporal} and then translating to average channel gains and residual bandwidth \cite{Guo2018Exploiting}.
\subsubsection{Optimizing Predictive Resource Allocation Plan}
We aim to optimize a resource allocation plan that minimizes the total transmission time required to ensure the quality of service (QoS) of each MS. The plan for MS$_k$ is denoted as ${\bf s}^k=[s_1^k,\cdots,s_{T_f}^k]^{\rm H}$, where $s_j^k$ is the fraction of time slots assigned to the MS in the $j$th frame. The objective function can be expressed as $\sum_{k=1}^K\sum_{j=1}^{T_f} s_j^k$. To guarantee the QoS, the requested file should be completely downloaded to the MS before an expected deadline. For mathematical simplicity, we let the duration between the time instant when a MS initiates a request and the transmission deadline equals to the duration of the prediction window. Then, the QoS constraint can be expressed as $\sum_{j=1}^{T_f}  s^k_{j} R^k_j/B_k\Delta=1$. Denote $r_j^k \triangleq R^k_j/B_k\Delta$ and ${\bf r}^k=[r_1^k,\cdots,r_{T_f}^k]^{\rm H}$, which is called \emph{average rate} in the sequel.
The optimization problem can be formulated as,
\begin{subequations}\label{P1-vector}
	\vspace{-1mm}
	\begin{align}
	{\bf P1}:\min_{\bf S} &~~~ \|{\bf S}\|_1 \\
	{\rm s.t.} &~~~{\bf S}\cdot{\bf R}^{\rm H}\star{\bf I}={\bf I},\label{P1-c-vec}\\
	&~~~{\bf S}\cdot{\bf M}_i^{\rm H}\star{\bf I} \preceq {\bf I}, i=1,\cdots,N_{\rm b},\label{P1-d-vec}
	\end{align}
\end{subequations}
where ${\bf S} = [{\bf s}^1,\cdots, {\bf s}^K], {\bf R} = [{\bf r}^1,\cdots, {\bf r}^K]$,  $({\bf M}_i)_{kj}=1$ or 0 if MS$_k$ associates or not associates to the $i$th BS in the $j$th frame, ${\bf I}$ is the identity matrix, $(\cdot)_{kj}$ stands for the element in the $k$th row and $j$th column of a matrix.  \eqref{P1-c-vec} is the QoS constraint, and \eqref{P1-d-vec} is the resource constraint that ensures the total time allocated in each frame of each BS  not exceeding one frame duration. In  \eqref{P1-c-vec} and \eqref{P1-d-vec}, ``$\cdot$'' denotes matrix multiplication, and ``$\star$'' denotes element wise multiplication, ${\bf A}\preceq{\bf B}$ and ${\bf A}\succeq{\bf B}$ mean that each element in ${\bf A}$ is not larger or smaller than each element in ${\bf B}$, respectively.

After the plan for each MS is made by solving \textbf{P1} at the start of the prediction window, a transmission progress can be computed according to the plan as well as the predicted average rates, which  determines how much data should be transmitted to each MS in each frame. Then, each BS schedules the MSs in its cell in each time slot, see details in \cite{YTCOM2016}.

\subsection{Unsupervised Learning for Resource Allocation Plan}
Problem \textbf{P1} is a convex optimization problem, which can be solved by interior-point method. However, the computational complexity scales with ${\cal O}(KT_f)^{3.5}$, which is prohibitive. To reduce on-line complexity, we can design a DNN to learn the optimal resource allocation plan. To reduce the off-line complexity in generating labels, we train the DNN with unsupervised learning. To this end, we transform problem \textbf{P1} into a functional optimization problem as suggested in \cite{sun2019pimrc}, from which the relationship between the optimal solution of \textbf{P1} and the known parameters can be found.

Denote the relationship as ${\bf S}({\bm \theta})$, which can be found from the following problem as proved in \cite{sun2019pimrc},
\vspace{-1mm}
\begin{subequations}\label{P2-vector}
	\begin{align}
	{\bf P2}:\min_{\bf S({\bm \theta})} &~~~ {\mathbb E}_{\bm \theta}\{\|{\bf S}({\bm \theta})\|_1\} \\
	{\rm s.t.} &~~~{\bf S({\bm \theta})}\cdot{\bf R}^{\rm H}\star{\bf I}={\bf I},\label{P2-c-vec}\\
	&~~~{\bf S({\bm \theta})}\cdot{\bf M}_i^{\rm H}\star{\bf I} \preceq {\bf I}, i=1,\cdots,N_{\rm b}.\label{P2-d-vec}
	\end{align}
\end{subequations}
where ${\bm \theta} = \{{\bf R}, {\bf M}_1, \cdots, {\bf M}_{N_{\rm b}}\}$ are the known parameters.

Problem \textbf{P2} is convex, hence it is equivalent to its Lagrangian dual problem \cite{convexopt},
\vspace{-1mm}
\begin{subequations}\label{P3}
	\begin{align}
	{\bf P3}:\notag\\\max_{\bm \lambda(\bm \theta)}\min_{\bf S({\bm \theta})} &{\cal L}\triangleq {\mathbb E}_{\bm \theta}\{\|{\bf S}({\bm \theta})\|_1\!+\!{\bm\mu}^{\rm H}({\bm \theta})\big({\bf S({\bm \theta})}\cdot{\bf R}^{\rm H}\star{\bf I}-{\bf I}\big)\!\cdot\!{\bm 1}+\notag\\
	& ~~~~~~\textstyle\sum_{i=1}^{N_{\rm b}}{\bm\nu}_i^{\rm H}({\bm \theta})\big({\bf S({\bm \theta})}\cdot{\bf M}_i^{\rm H}\star{\bf I}- {\bf I}\big)\!\cdot\!{\bm 1}\}\label{P3-o} \\
	{\rm s.t.} &~~~{\bm\nu}^i(\bm \theta)\succeq {\bm 0}, \forall i\in\{1,\cdots,N_{\rm b}\},\label{P3-c-vec}
	\end{align}
\end{subequations}
where ${\cal L}$ is the Lagrangian function, ${\bm\lambda}^{\rm H}(\bm \theta)=[{\bm\mu}^{\rm H}(\bm \theta), {\bm\nu}_1^{\rm H}(\bm \theta),\cdots, {\bm\nu}_{N_{\rm b}}^{\rm H}(\bm \theta)]$ is the vector of Lagrangian multipliers. Considering the universal approximation theorem \cite{Hornik1989UnivApprox}, ${\bf S}({\bm \theta})$ and ${\bm\lambda}(\bm \theta)$ can be approximated with DNN  \cite{sun2019pimrc}.

\subsubsection{Design of the DNN} \label{sec: DNN design}
The input of a DNN to learn ${\bf S}({\bm \theta})$ can be designed straightforwardly as ${\bm \theta} = \{{\bf R}, {\bf M}_1, \cdots, {\bf M}_{N_{\rm b}}\}$, which is of high dimension. To reduce the training complexity, we learn the resource allocated by each BS (say the $i$th BS) by a neural network called DNN-$s$.

The input of DNN-$s$ is ${\bf x}_i={\rm vec}({\bf R} \star {\bf M}_i)=[({\bf x}_i^1)^{\rm H}, \cdots, ({\bf x}_i^{K_{\max}})^{\rm H}]^{\rm H}$, where ${\rm vec}(\cdot)$ denotes the operation of concatenating each column of a matrix into a vector, ${\bf x}_i^k$ is the average rate of MS$_k$ if it served by the $i$th BS. If $K<K_{\max}$, then ${\bf x}_i^k={\bf 0}$ for $\forall k>K$.
The output of DNN-$s$ is the resource allocation plan of all the MSs when they are served by the $i$th BS, which is normalized by the total resources allocated to each MS to meet the constraint in \eqref{P2-c-vec}, i.e.,
$
\textstyle\hat{s}_j^k\!=\!\frac{\hat{s}_j^{k'} r_j^k}{\sum_{\tau=1}^{T_f}\hat{s}_{\tau}^{k'}r_{\tau}^k} \Big/r_j^k\!=\! \frac{\hat{s}_j^{k'}}{\sum_{\tau=1}^{T_f}\hat{s}_{\tau}^{k'}r_{\tau}^k},k\!=\!1,\!\cdots\!,K, j\!=\!1,\!\cdots\!,T_f$,
where $\hat{s}_j^{k'}$ and $\hat{s}_j^k$ are respectively the output of DNN-$s$ before and after normalization.
We use the commonly used \texttt{Softplus} (i.e., $y=g(x)\triangleq\log(1+\exp(x))$) as the activation function of the hidden layers and output layer to ensure the learned plan being equal or larger than $0$. Since DNN-$s$ is used to learn ${\bf S}({\bm \theta})$ that is permutation invariant to ${\bf x}_i$, we can apply parameter sharing introduced in section \ref{sec: param share} to learn its input-output relationship $f_s({\bf x}_i, {\bf W}_s)$, where ${\bf W}_s$ denotes all the weight matrices and biases in DNN$-s$.

To learn the Lagrange multipliers, we design a fully-connected neural network called DNN-$\lambda$, which has different structure from  DNN-$s$ and cannot use parameter sharing. Since the constraint in \eqref{P2-c-vec} already holds due to the normalization operation in the output of DNN-$s$, we do not need to learn multiplier ${\bm \mu}$ in \eqref{P3-o} and hence we only learn multiplier ${\bm \nu}_i$. Since ${\bm \nu}_i$ is used to satisfy constraint \eqref{P2-d-vec}, which depends on ${\bf x}_i$, the input of DNN-$\lambda$ is the same as DNN-$s$. The activation functions in hidden layers and output layer are also \texttt{Softplus} to ensure the Lagrange multipliers being equal or larger than $0$, hence \eqref{P3-c-vec} can be satisfied. The input-output relationship of DNN-$\lambda$ is denoted as $f_{\nu}({\bf x}_i, {\bf W}_{\nu})$.

\subsubsection{Training Phase}

DNN-$s$ and DNN-$\lambda$ are trained in multiple epochs, where in each epoch ${\bf W}_s$ and ${\bf W}_{\nu}$ are sequentially updated using the gradients of a cost function with respective to ${\bf W}_s$ and ${\bf W}_{\nu}$ via back-propagation. The cost function is the empirical form of \eqref{P3-o}, where ${\bf S}({\bm \theta})$ and ${\bm \lambda}({\bm \theta})$ are replaced by $f_s({\bf x}_i, {\bf W}_s)$ and $f_{\nu}({\bf x}_i, {\bf W}_{\nu})$. In particular, we replace ${\mathbb E}_{\bm\theta}\{\cdot\}$ in the cost function with empirical mean, because the probability density function of ${\bm\theta}$ is unknown. We omit the second term in \eqref{P3-o} because the constraint in \eqref{P2-c-vec} can be ensured by the normalization operation in the output of DNN-$s$. The cost function is expressed as,

\vspace{-5mm}
\begin{small}
\begin{eqnarray}
&&\!\!\!\!\!\!\!\!\hat{\cal L}({\bf W}_s, {\bf W}_{\nu}) \notag\\
&\!\!\!\!\!\!\!\!\!\!\!\!\!\!\!\!\!\!\!\!=\!\!\!\!\!\!\!\!\!\!&\!\!\!\!\!\!\!\!\!\! \frac{1}{N}\! \sum_{n=1}^N \sum_{i=1}^{N_{\rm b}}\left( \|{\bm f}_{s,i}^{(n)}\|_1 \!+\! ({\bm f}_{\nu,i}^{(n)})^{\rm H}\big( [{\bm f}_{s,i}^{(n)}]_{T_f\times {K_{\max}}}\! \cdot \!{\bf M}_i^{\rm H} \!\star\! {\bf I} \!-\! {\bf I}\big)\!\cdot\!{\bm 1}\right)\!,\notag \nonumber
\end{eqnarray}
\end{small}where ${\bm f}_{s,i}^{(n)}\triangleq f_s({\bf x}_i^{(n)}, {\bf W}_s)$, ${\bm f}_{\nu,i}^{(n)}\triangleq f_{\nu}({\bf x}_i^{(n)}, {\bf W}_{\nu})$, and ${\bf x}^{(n)}_i$ denotes the $n$th sample, $[{\bf a}]_{m\times n}$ is the operation to represent vector ${\bf a}$ as a matrix with $m$ rows and $n$ columns. In DNN-$s$, ${\bf W}_s$ is trained to minimize $\hat{\cal L}({\bf W}_s, {\bf W}_{\nu})$. In DNN-$\lambda$, ${\bf W}_{\nu}$ is trained to maximize $\hat{\cal L}({\bf W}_s, {\bf W}_{\nu})$. The learning rate is adaptively updated with Adam algorithm \cite{Kingma2014Adam}.
\subsubsection{Operation Phase}
For illustration, assume that $\bf R$ and ${\bf M}_i, i=1,\cdots,N_{\rm b}$ are known at the beginning of the prediction window. Then, by sequentially inputting the trained DNN-$s$ with ${\bf x}_i={\rm vec}({\bf R} \star {\bf M}_i), i=1,\cdots,N_{\rm b}$, DNN-$s$ can sequentially output the resource allocation plans for all MSs served by the $1,\cdots,N_{\rm b}$th BS.

\section{Simulation Results}
In this section, we evaluate the performance of the proposed solution with simulations.

Consider a cellular network with cell radius $R_{\rm b}=250$ m, where four BSs each equipped with $N_{\rm tx}=8$ antennas are located along a straight line. For each BS, $P_{\max}$ is 40 W, $W_{\max}=20$ MHz and the cell-edge SNR is set as 5 dB, where the intercell interference is implicitly reflected. The path loss model is $36.8 + 36.7\log_{10}(d)$, where $d$ is the distance between the BS and MS in meter. The MSs move along three roads of straight lines with minimum distance from the BSs as $50$ m, $100$ m and $150$ m, respectively. At the beginning of the prediction window, $K$ MSs each requests a file with size of $B_k=1$ Mbytes (MB).
Each frame is with duration of $\Delta=1$ second, and each time slot is with duration $10$ ms, i.e., each frame contains $T_s=100$ time slots.

To characterize the different resource usage status of the BSs by serving the RT traffic in an under-utilized network, we consider two types of BSs:  busy BS  with average residual bandwidth in the prediction window $\overline{W}= 5$ MHz and idle BS with $\overline{W}= 10$ MHz, which are alternately located along the line as idle, busy, idle, busy. The residual transmit power is proportional to the residual bandwidth \cite{YTCOM2016}. The results are obtained from 100 Monte Carlo trials. In each trial, $K$ is randomly selected from 1 to $K_{\max}=40$, the MSs initiate requests randomly at a location along the trajectory, and the trajectories change randomly with speed uniformly distributed in $(10, 25)$ m/s and directions uniformly selected from 0 or +180 degree. The small-scale channel in each time slot changes independently according to Rayleigh fading, and the residual bandwidth at each BS in each time slot varies according to Gaussian distribution with mean value $\overline{W}$ and standard derivation $0.2\overline{W}$. The setup is used in the sequel unless otherwise specified.

The fine-tuned hyper-parameters for the DNNs when $T_f=30$ seconds are summarized in Table \ref{table}. When $T_f$ changes, the hyper-parameters should be tuned again to achieve the best performance. The training set contains 10,000 samples and the testing set contains 100 samples. In order to increase the generalization ability on the number of MSs, the training samples are generated when $K=K_{\max}$.

\begin{table}[htb!]
	\centering
	\vspace{-2mm}
	\caption{Hyper-parameters for the Two DNNs When $T_f=30$ Seconds}\label{table}
	\vspace{-2mm}
	\footnotesize
	\begin{tabular}{c|c|c}
		\hline\hline
		\multirow{2}{*}{\textbf{Parameters}} & \multicolumn{2}{c}{\textbf{Values}} \\
		\cline{2-3}
		~ & DNN-$s$ & DNN-$\lambda$  \\
		\hline
		\tabincell{c}{Number of input nodes} & $K_{\max}T_f = 600$ & $K_{\max}T_f = 600$ \\
		\hline
		\tabincell{c}{Number of hidden layers} & 2 & 2 \\
		\hline
		\tabincell{c}{Number of hidden nodes} & 1000, 1000 & 200, 100 \\
		\hline
		\tabincell{c}{Number of output nodes} &  $K_{\max}T_f = 600$ &  $T_f = 30$ \\
		\hline
		\tabincell{c}{Initial learning  rate} & \multicolumn{2}{c}{0.01} \\
		\hline\hline
	\end{tabular}
\vspace{-5mm}
\end{table}

%
%
%

\subsection{Sample and Computational Complexity}

Sample complexity is defined as the minimal number of training samples for a DNN to achieve an expected performance. We provide the sample complexity when the objective in \textbf{P1} on the testing set can achieve less than 20\% performance loss from the optimal value (i.e., the total allocated time resource for all MSs), which is obtained by solving \textbf{P1} with interior-point method. Both DNN-$s$ with and without parameter sharing are trained with $200$ epochs. The results are shown in Table \ref{table: parameter sharing}. As expected, the sample complexity of the DNN-$s$ with parameter sharing is much less than the DNN-$s$ without parameter sharing.

\begin{table}[htb!]
	\centering
	\vspace{-2mm}
	\caption{Sample Complexities of DNNs}\label{table: parameter sharing}
	\vspace{-2mm}
	\footnotesize
	\begin{tabular}{c|c|c|c}
		\hline\hline
		~ & $T_f=5$ s & $T_f=15$ s & $T_f=30$ s \\
		\hline
		With parameter sharing & 2,000 & 3,000 & 4,000 \\
		\hline
		Without parameter sharing & 7,000 & 10,000 & 10,000 \\
		\hline\hline
	\end{tabular}
\vspace{-3mm}
\end{table}

To show how much computational complexity can be reduced by parameter sharing, we train two groups of DNNs, (i) DNN-$s$ with parameter sharing and DNN-$\lambda$, (ii) DNN-$s$ without parameter sharing and DNN-$\lambda$, with same number of training samples. Since the two groups of DNNs have the same number of layers, same number of neurons in each layer and are trained with same number of samples, the computational complexity in each epoch is identical for the two groups of DNNs. Then, the computational complexity in the training phase can be measured by the number of training epochs required to achieve the expected performance. The results when $B_k=1$ MB and $B_k\thicksim\mathbb U(1,3)$ MB are shown in Table \ref{table: computation complexity}, where each DNN-$s$ contains three hidden layers each with 1000 neurons and is trained with 16000 samples  when $B_k\thicksim\mathbb U(1,3)$ MB.

\begin{table}[htb!]
	\centering
	\vspace{-2mm}
	\caption{Computational Complexities of DNNs}\label{table: computation complexity}
	\vspace{-2mm}
	\footnotesize
	\begin{tabular}{c|c|c|c|c}
		\hline\hline
		$B_k$ & ~& $T_f=5$ s & $T_f=15$ s & $T_f=30$ s \\
		\hline
		\multirow{2}{*}{1 MB} & w sharing & 20 epochs & 33 epochs & 32 epochs \\
		\cline{2-5}
		~ & w/o sharing & 59 epochs & 80 epochs & 60 epochs \\
		\hline
		\multirow{2}{*}{$\mathbb U(1,3)$ MB} & w sharing & 19 epochs & 24 epochs & 35 epochs \\
		\cline{2-5}
		~ & w/o sharing & 97 epochs & 101 epochs & 170 epochs \\
		\hline\hline
	\end{tabular}
\vspace{-3mm}
\end{table}

It is shown that the computational complexity of the DNN with parameter sharing is much less than the DNN without parameter sharing, especially for the case where $B_k\thicksim {\mathbb U}(1,3)$ MB. This is because in this case the wireless network is extremely busy, hence it is hard for DNN to learn a resource allocation plan that satisfies constraint \eqref{P2-c-vec}.

It should be noted that although we do not apply parameter sharing in DNN-$\lambda$, The convergence speed in the training phase is still faster by only applying parameter sharing to DNN-$s$. This is because the fine-tuned DNN-$\lambda$ has much simpler structure as shown in Table \ref{table}.

\subsection{Performance of PRA Learned with DNNs}
To evaluate the performance of the proposed method using the DNN with parameter sharing and unsupervised learning (with legend ``\textbf{Proposed}''), we compare the total transmission time required for downloading the files averaged over all MSs with the following methods.
\begin{itemize}
	\item \textbf{Supervised}: This is the method where the resource allocation plan is obtained by a DNN with parameter sharing trained in the supervised manner, where the labels in the training samples are generated by solving \textbf{P1} with interior-point method.
	\item \textbf{Optimal}: This is the optimal PRA policy, where the resource allocation plan is obtained by solving problem \textbf{P1} with interior-point method.
	\item \textbf{Baseline}: This is a non-predictive method \cite{su2015user}, where each BS serves the MS with the earliest deadline in each time slot. If several MSs have the same deadline, then the MS with most bits to be transmitted is served firstly.
\end{itemize}

\begin{figure}[!htb]
	\centering
	\vspace{-2mm}
	\includegraphics[width=0.75\linewidth]{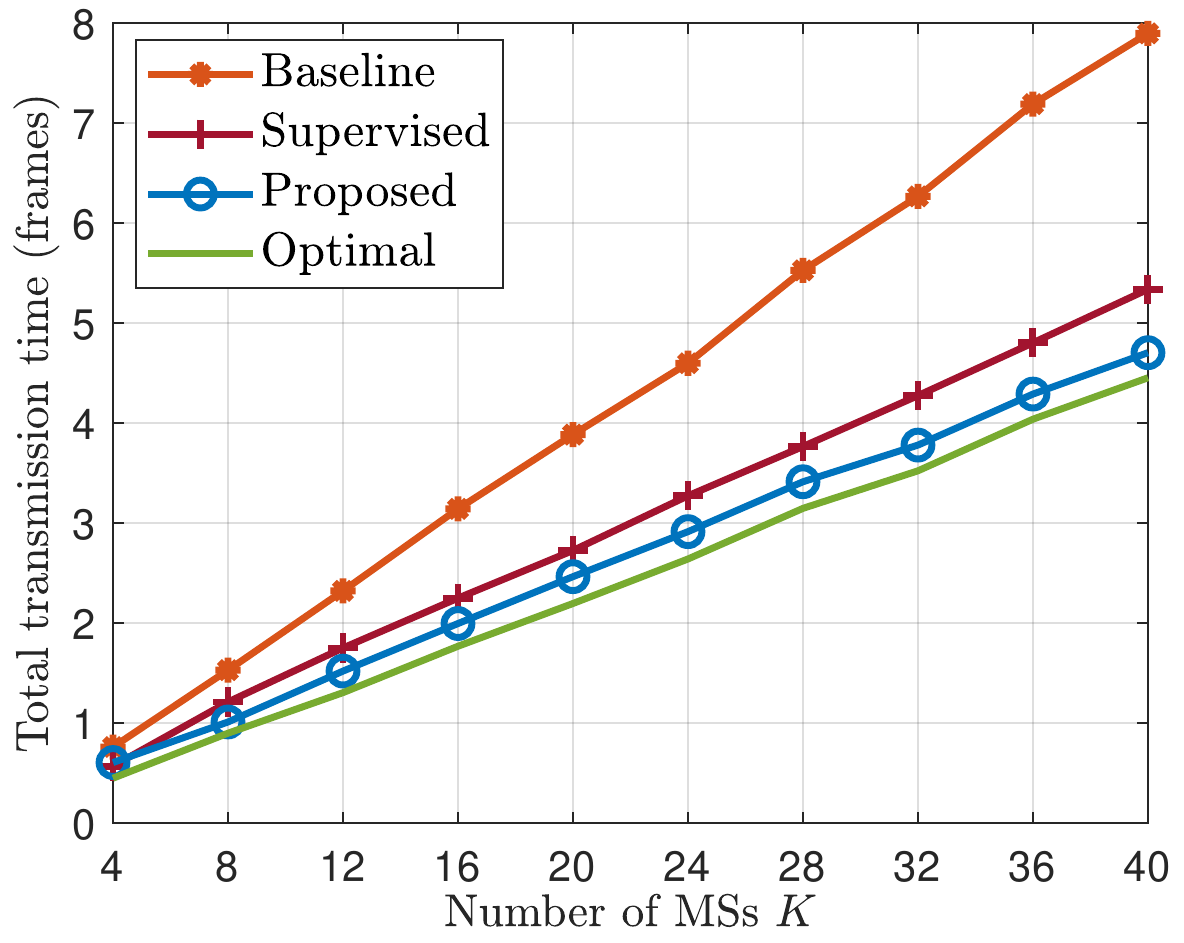}
	\vspace{-3mm}
	\caption{Performance comparison of all methods, $T_f=30$ seconds.}
	\vspace{-5mm}
	\label{fig:figtimearr}
\end{figure}

The results are shown in Fig. \ref{fig:figtimearr}. We can see that the proposed method performs closely to the optimal policy and dramatically outperforms the baseline. Besides, the proposed method with unsupervised DNN outperforms the method with supervised DNN. This is because the resource allocation policy learned from labels cannot satisfy the constraints in problem \textbf{P1}, which leads to resource confliction among users.
\section{Conclusions} \label{sec: conclusion}
In this paper, we proposed a structure of DNN with parameter sharing, which exploits \emph{a priori} information in many wireless tasks, i.e., permutation invariant. We used a case study to illustrate how this structure can be applied, where unsupervised DNN is used to learn the optimal solution of predictive resource allocation. Simulation results showed that the proposed DNN with parameter sharing performs closely to the numerically obtained optimal solution, and can reduce the sample and computational complexity for training remarkably compared to the fully-connected DNN without parameter sharing.
\vspace{-1mm}
\begin{appendices}
\numberwithin{equation}{section}
\vspace{-1mm}
\section{Proof of proposition \ref{pp: 1}}\label{appendix: A}
\vspace{-1mm}
We first prove the necessity. Assume that the function $f({\bf x})$ is permutation invariant to ${\bf x}$. If the $k$th block ${\bf x}_k$ in ${\bf x}=[{\bf x}^1, \cdots, {\bf x}^K]$ is changed to another position in ${\bf x}$ while the permutation of other blocks in ${\bf x}$ remains unchanged, i.e.,
\begin{equation}
\textstyle\tilde{\bf x}=[\underbrace{{\bf x}_1,\cdots,{\bf x}_{k-1}}_{(a)},\underbrace{{\bf x}_{k+1},\cdots,{\bf x}_K}_{(b)}],
\end{equation}
 where ${\bf x}_k$ may be in the blocks in $(a)$ or $(b)$, then
 $\tilde{\bf y}_k={\bf y}_{k-1}$ if ${\bf x}_k$ is in $(a)$ and $\tilde{\bf y}_k={\bf y}_{k+1}$ if ${\bf x}_k$ is in $(b)$, hence $\tilde{\bf y}_k\neq {\bf y}_k$. This indicates that the $k$th output block should change with the $k$th input block ${\bf x}_k$. On the other hand, if the position of ${\bf x}_k$ remains unchanged while the positions of other blocks ${\bf x}$ arbitrarily change, i.e., $\tilde{\bf x}=[{\bf x}^{N_1},\cdots,{\bf x}^{N_k-1},{\bf x}^k,{\bf x}^{N_{k+1}},\cdots,{\bf x}^{N_k}]$, then $\tilde{\bf y}=[{\bf y}^{N_1},\cdots,{\bf y}^{N_k-1},{\bf y}^k,{\bf y}^{N_{k+1}},\cdots,{\bf x}^{N_k}]$, and $\tilde{\bf y}_k={\bf y}_k$. This means that ${\bf y}_k$ is not affected by the permutation of the input blocks other than ${\bf x}_k$. Therefore, the function should have the form in \eqref{eq: perm inva}.

We then prove the sufficiency. Assume that the function $f({\bf x})$  has the form in \eqref{eq: perm inva}. If ${\bf x}$ is changed to $\tilde{\bf x}=[{\bf x}^{N_1},\cdots, {\bf x}^{N_K}]$, then the $k$th block of $\tilde{\bf y}$ is $\tilde{\bf y}^k = \zeta(\psi({\bf x}^{N_k}),{\cal F}_{n=1,n\neq N_k} \phi({\bf x}^n))={\bf y}^{N_k}$. Hence, the output corresponding to $\tilde{\bf x}$ is $\tilde{\bf y}=[\tilde{\bf y}^1,\cdots,\tilde{\bf y}^K]=[{\bf y}^{N_1},\cdots,{\bf y}^{N_K}]$. According to Definition \ref{def: 1}, the function in \eqref{eq: perm inva} is permutation invariant to ${\bf x}$.
\vspace{-2mm}

\section{Proof of proposition \ref{pp: 2}}\label{appendix: B}
\vspace{-1mm}
For notational simplicity, in the following proof we omit the biases in the DNN. With the weight matrices in \eqref{weight mat0}, the output of the 2nd hidden layer is ${\bf h}^{[2]} = g({\bf W}^{[1,2]}{\bf x})$, and the output of the $l$th hidden layer $(2<l<L)$ can be written as,
\begin{eqnarray}\label{B1}
{\bf h}^{[l]} \!\!\!&\!\!\!\!=\!\!\!\!&\!\!\!g({\bf W}^{[l-1,l]}{\bf h}^{[l-1]}) \notag\\
\!\!\!&\!\!\!\!=\!\!\!\!&\!\!\!\textstyle\big[g\big({\bf U}^{[l-1,l]}{\bf h}^{[l-1],1}\!+\!{\bf V}^{[l-1,l]}\sum_{k=2}^K {\bf h}^{[l-1],k}\big),\cdots,\notag\\
&&\textstyle g\big({\bf U}^{[l-1,l]}{\bf h}^{[l-1],K}\!+\!{\bf V}^{[l-1,l]}\sum_{k=1}^{K-1} {\bf h}^{[l-1],k}\big)\big],
\end{eqnarray}
where ${\bf h}^{[l]}=[{\bf h}^{[l],1},\cdots,{\bf h}^{[l],K}]$. From \eqref{B1} the $k$th block of ${\bf h}^{[l]}$ can be expressed as ${\bf h}^{[l],k}\!\!\!=\!\!\!g\big({\bf U}^{[l-1,l]}{\bf h}^{[l-1],k}\!+\!{\bf V}^{[l-1,l]}\sum_{n=1,n\neq k}^K {\bf h}^{[l-1],n}\big)$.

We can see that the relation between ${\bf h}^{[l],k}$ and ${\bf h}^{[l-1],k}$ has the same form as in \eqref{eq: perm inva}. Hence $g({\bf W}^{[l-1,l]}{\bf h}^{[l-1]})$ is permutation invariant to ${\bf h}^{[l-1]}$.
Since the output of every hidden layer is permutation invariant to the output of its previous layer, and ${\bf y} =g({\bf W}^{[L-1,L]}{\bf h}^{[L-1]})$ is also permutation invariant to ${\bf h}^{[L-1]}$, $f({\bf x}, {\bf W})$ in \eqref{eq: FC func} is permutation invariant to ${\bf x}$.
\end{appendices}
\vspace{-1mm}
	\bibliography{IEEEabrv,GJ1}
	\vspace{-1mm}
\end{document}